\documentclass{article}

\usepackage{microtype}
\usepackage{graphicx}
\usepackage{subcaption}
\usepackage{booktabs} 
\usepackage{multirow} 

\usepackage{hyperref}

\usepackage[preprint]{icml2026}

\usepackage{amsmath}
\usepackage{amssymb}
\usepackage{mathtools}
\usepackage{amsthm}
\usepackage[table]{xcolor}

\usepackage[capitalize,noabbrev]{cleveref}

\theoremstyle{plain}

\theoremstyle{definition}

\theoremstyle{remark}

\definecolor{highlightblue}{RGB}{230, 240, 255}

\usepackage[textsize=tiny]{todonotes}

\icmltitlerunning{Mitigating Premature Discretization with Progressive Quantization
for Robust Vector Tokenization}

\begin{document}

\twocolumn[
  \icmltitle{Mitigating Premature Discretization with Progressive Quantization \\ 
  for Robust Vector Tokenization}



  \icmlsetsymbol{equal}{*}

  \begin{icmlauthorlist}
    \icmlauthor{Wenhao Zhao}{equal,yyy}
    \icmlauthor{Qiran Zou}{equal,yyy}
    \icmlauthor{Zhouhan Lin}{comp}
    \icmlauthor{Dianbo Liu}{yyy}
  \end{icmlauthorlist}

  \icmlaffiliation{yyy}{National University of Singapore, Singapore}
  \icmlaffiliation{comp}{Shanghai Jiao Tong University, Shanghai, China}

  \icmlcorrespondingauthor{Wenhao Zhao}{e1374536@u.nus.edu}
  \icmlcorrespondingauthor{Dianbo Liu}{dianbo@nus.edu.sg}

  \icmlkeywords{Machine Learning, ICML}

  \vskip 0.3in
]



\printAffiliationsAndNotice{}  

\begin{abstract}
 Vector Quantization (VQ) has become the cornerstone of tokenization for many multimodal Large Language Models and diffusion synthesis. However, existing VQ paradigms suffer from a fundamental conflict: they enforce discretization before the encoder has captured the underlying data manifold. We term this phenomenon \textit{Premature Discretization}. To resolve this, we propose Progressive Quantization (\texttt{ProVQ}), which incorporates the dynamics of quantization hardness as a fundamental yet previously overlooked axis in VQ training. By treating quantization as a curriculum that smoothly anneals from a continuous latent space to a discrete one, \texttt{ProVQ} effectively guides the codebook toward the well-expanded manifolds. Extensive experimental results demonstrate the broad effectiveness of \texttt{ProVQ} across diverse modalities. We report improved reconstruction and generative performance on the ImageNet-1K and ImageNet-100 benchmarks, highlighting the \texttt{ProVQ}'s boost for generative modeling. Furthermore, \texttt{ProVQ} proves highly effective for modeling complex biological sequences, establishing a new performance ceiling for protein structure tokenization on the StrutTokenBench leaderboard.
 
\end{abstract}

\section{Introduction}

Vector Quantization (VQ)\cite{van2017neural} has emerged as a fundamental bridge between raw continuous signals and the discrete symbolic processing required by modern generative models. By mapping high-dimensional data into a finite set of learnable codebook vectors, VQ serves as the cornerstone for scaling Large Language Models (LLMs) to multimodal domains\cite{chang2022maskgit,gao2025foldtoken,dhariwal2020jukebox, esser2021taming}, powers the latent spaces of high-fidelity Diffusion Models\cite{gu2022vectordiffusion, tang2022improveddiffusion}, and provides the compressed representations necessary for complex signal synthesis. However, despite its ubiquity, training stable VQ-based models remains a notorious challenge, often necessitating sensitive hyperparameter tuning or heuristic-driven interventions\cite{huh2023straightening}.

\begin{figure}[t]
  \begin{center}
    \centerline{\includegraphics[width=\columnwidth]{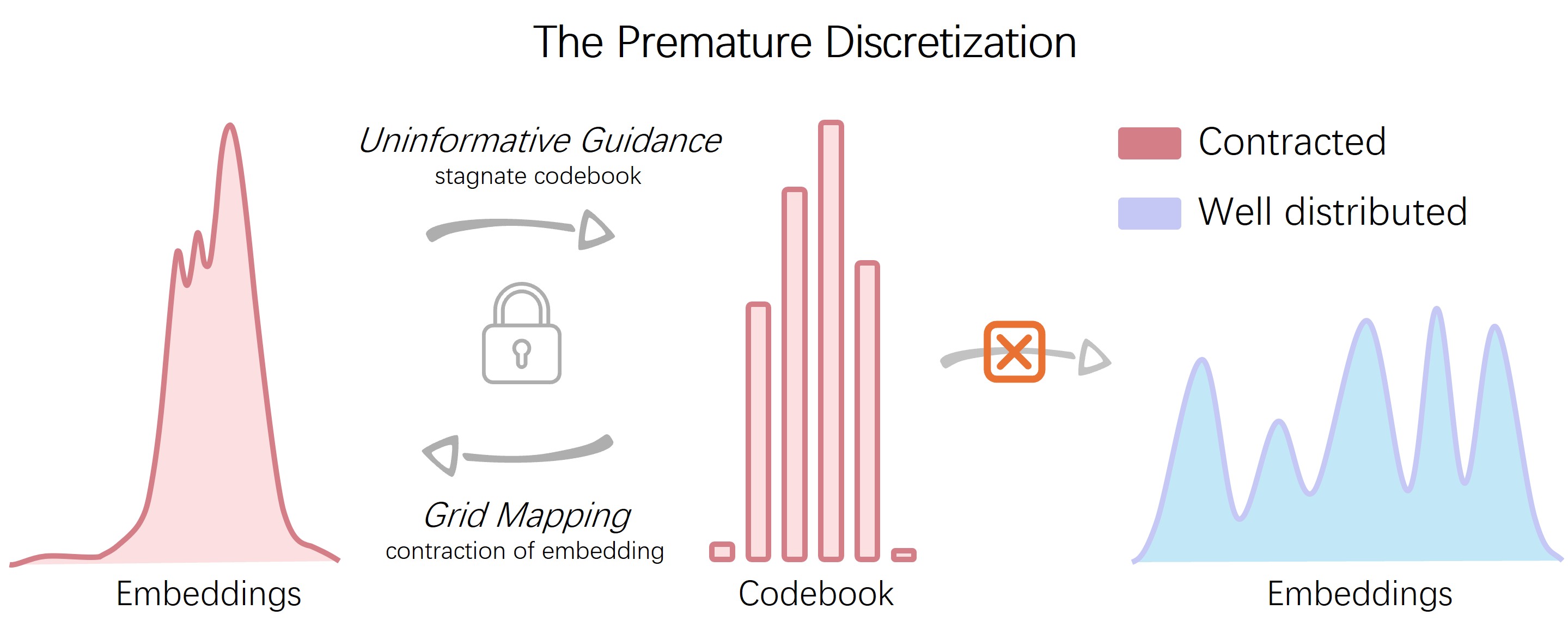}}
    \caption{\textbf{The \textit{Premature Discretization} and resulting optimization deadlock.} During early training stages, grid mapping forces the embedding distribution to contract and align with a sub-optimal clustered code, while uninformative guidance of embeddings causes the codebook vectors to stagnate. This mutual constraint creates a rigid optimization deadlock, which traps the model in a local minimal state and prevents it from exploring the full, well-distributed latent manifold (right).}
    \label{fig_intro}
  \end{center}
  \vspace{-1.25em}
\end{figure}

In this paper, we analyse a fundamental optimization bottleneck in standard VQ training which we term \textit{Premature Discretization}. At the onset of training, both the encoder and the codebook are initialized randomly, creating a destructive ``chicken-and-egg'' cycle that leads to a co-adaptation deadlock. Specifically, the encoder requires a meaningful codebook to provide stable gradient signals for manifold learning, while the codebook conversely depends on consistent, well-clustered encoder outputs to optimize its representative centroids. As illustrated in Figure~\ref{fig_intro}, when a hard discrete bottleneck is enforced prematurely, the model is forced into a reciprocal failure of representation where the learning process is stagnated.

This deadlock manifests through two simultaneous and reinforcing phenomena. First, grid mapping forces the encoder's embedding distribution to prematurely contract and align with a sub-optimal random grid. Simultaneously, uninformative embeddings cause codebook vectors to stagnate. Consequently, this mutual constraint creates an optimization deadlock that traps the model in a sub-optimal state, thereby preventing the encoder and codebook from exploring the full, well-distributed latent manifold.

We hypothesize the core issue of this premature coupling is that it entirely halts the manifold warmup phase. Because the encoder and codebook co-adapt to unlearned noise rather than the underlying data distribution, gradient fluidity is polluted during the critical early stages of training. It makes the resulting latent space poorly organized, ultimately failing to capture the expressive modes necessary for high-fidelity synthesis. In this study, we conduct a systematic analysis of this phenomenon and find that this phenomenon is a result of a structural and mechanistic conflict in the discrete representation learning process, in which the encoder and codebook have entered a destructive co-adaptation phase that leads to the encoder failing to ”unfold” the whole data manifold.  While existing literature has proposed various heuristics to repair the latent space after the fact, these methods generally treat the symptoms of poor utilization rather than the root cause.

To resolve this issue, we propose \textbf{Progressive Vector Quantization (\texttt{ProVQ})}. We frame VQ training as a curriculum learning\cite{bengio2009curriculum} problem to disentangle the continuous and discrete learning at early stage, where the model first masters the ``easy'' task of continuous manifold warmup before being challenged with the ``hard'' constraint of discrete quantization. By introducing a soft-to-hard transition axis, \texttt{ProVQ} maintains gradient fluidity, allowing the encoder to unfold the continuous data manifold in a stable environment. As the training progresses, these continuous representations are gradually compressed into discrete codes through a scheduled co-adaptation process, ensuring the final codebook is a refinement of an already optimized latent space.

Our contributions are summarized as follows:
\begin{itemize}
    \item We characterize how the co-adaptation between the encoder and codebook becomes trapped in sub-optimal local minima.
     \item We introduce a minimal synthetic diagnostic tool for revealing discretization pathologies.
    \item We introduce Progressive Vector Quantization (\texttt{ProVQ}), a curriculum-based training strategy designed to prevent premature stagnation by decoupling manifold warmup from latent discretization.
    \item We demonstrate \texttt{ProVQ} improves in reconstruction and generative performance on ImageNet-1K and ImageNet-100 over LlamaGen.
    \item We show that \texttt{ProVQ} is highly effective for protein structure modeling on StrutTokenBench, achieving the state-of-art performance.
\end{itemize}

\section{Related Works}

\vspace{-0.2cm}
The stability and utilization of discrete representation learning have long been central themes in the evolution of neural quantization. Our work situates itself at the intersection of quantization heuristics and curriculum learning, specifically addressing the dynamic relationship between the latent space and the codebook.

The Vector Quantized Variational Autoencoder (VQ-VAE) \cite{van2017neural} established the foundation for discrete bottlenecks by mapping encoder outputs to the nearest entry in a learnable codebook. Building on this, VQGAN \cite{esser2021taming} enhanced visual reconstruction quality through the integration of adversarial and perceptual losses. Subsequent research has proposed various methods to improve the robustness and representational capacity of the VQ-VAE framework. These include strategies like codebook restarts \cite{dhariwal2020jukebox}, where underutilized entries are re-initialized, and architectural constraints such as Factorized Codes \cite{yu2021vector}. Furthermore, SimVQ \cite{simvq} introduced a reparameterization of code vectors through a learnable linear transformation layer over a latent basis, aiming to simplify and improve the efficiency of codebook optimization.

The adoption of vector quantization has catalyzed progress across diverse domains by mapping continuous data into discrete modality. In computer vision side, the discretization of latent spaces allows generative models like LlamaGen\cite{llamagen} and VAR\cite{tian2024visual} to treat image synthesis as a sequence modeling task. This paradigm extends to structural biology, where tokenizing complex 3D protein topologies enables the use of protein language models\cite{ESM3, gao2025foldtoken}. Similarly, in audio field, vqvae has been widely used as codec including Soundstream\cite{zeghidour2021soundstream} and Wavtokenizer \cite{ji2024wavtokenizer} .

Our method is inspired by curriculum learning \cite{bengio2009curriculum}, which suggests that models learn better when the task complexity increases gradually. Similar concepts have appeared in Gumbel-Softmax annealing \cite{jang2016categorical}, which uses a temperature parameter to transition from a soft distribution to a one-hot encoding. While Gumbel-Softmax is widely used in categorical settings, applying a similar soft-to-hard logic directly to the geometry of the vector quantization space—specifically via a manifold warmup—remains underexplored. Our work frames the whole discretization process as the curriculum, boosting the optimization process of vector quantization.

\begin{figure*}[ht]
  \begin{center}
    \centerline{\includegraphics[width=2\columnwidth]{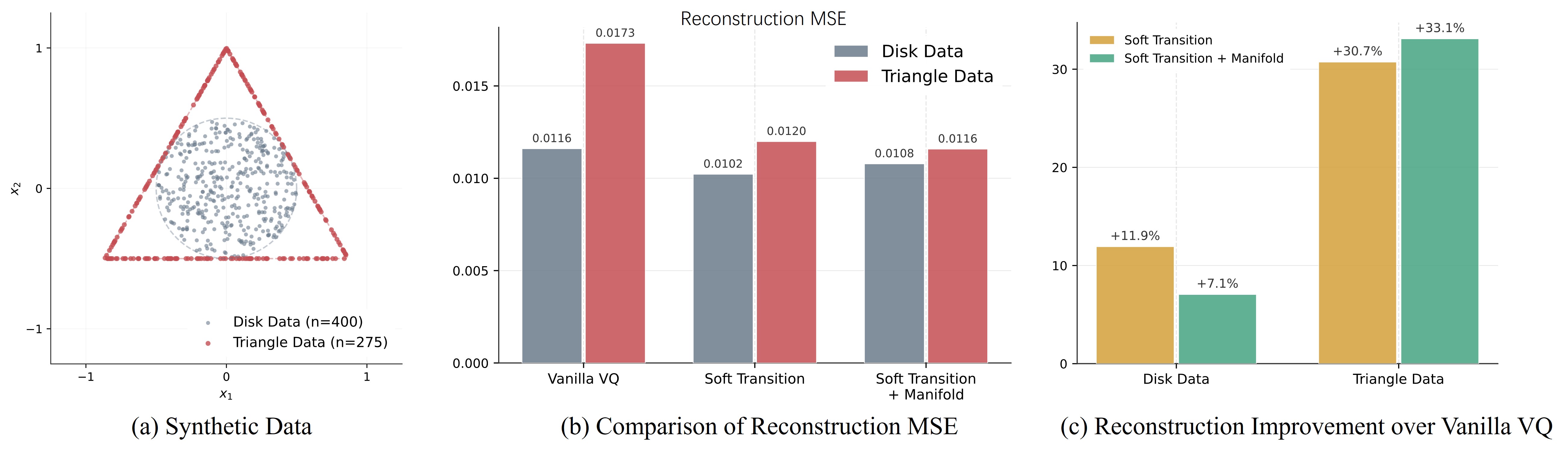}}
    \caption{\textbf{Empirical Validation on Synthetic 2D datasets.} 
    (a) Synthetic dataset composed by Disk shape data plus triangle data to make gridding mapping visible by edge of triangle. 
    (b) Comparison of reconstruction performance over different configurations, demonstrating that both the Soft Transition and the full \texttt{ProVQ} (Soft Transition + Manifold) strategies consistently outperform the Vanilla VQ baseline. 
    (c) Reconstruction improvement relative to Vanilla VQ. While a Soft Transition alone yields substantial gains ($+11.9\%$ for Disk and $+30.7\%$ for Triangle), the integration of a Manifold Warmup further boosts performance, achieving a $+33.1\%$ improvement on the triangle dataset. 
    These results underscore that decoupling continuous and discrete learning at early stage.}
    \label{fig_synthetic_exp_results_and_input}
  \end{center}
\end{figure*}

\begin{figure*}[h]
  \begin{center}
    \centerline{\includegraphics[width=2\columnwidth]{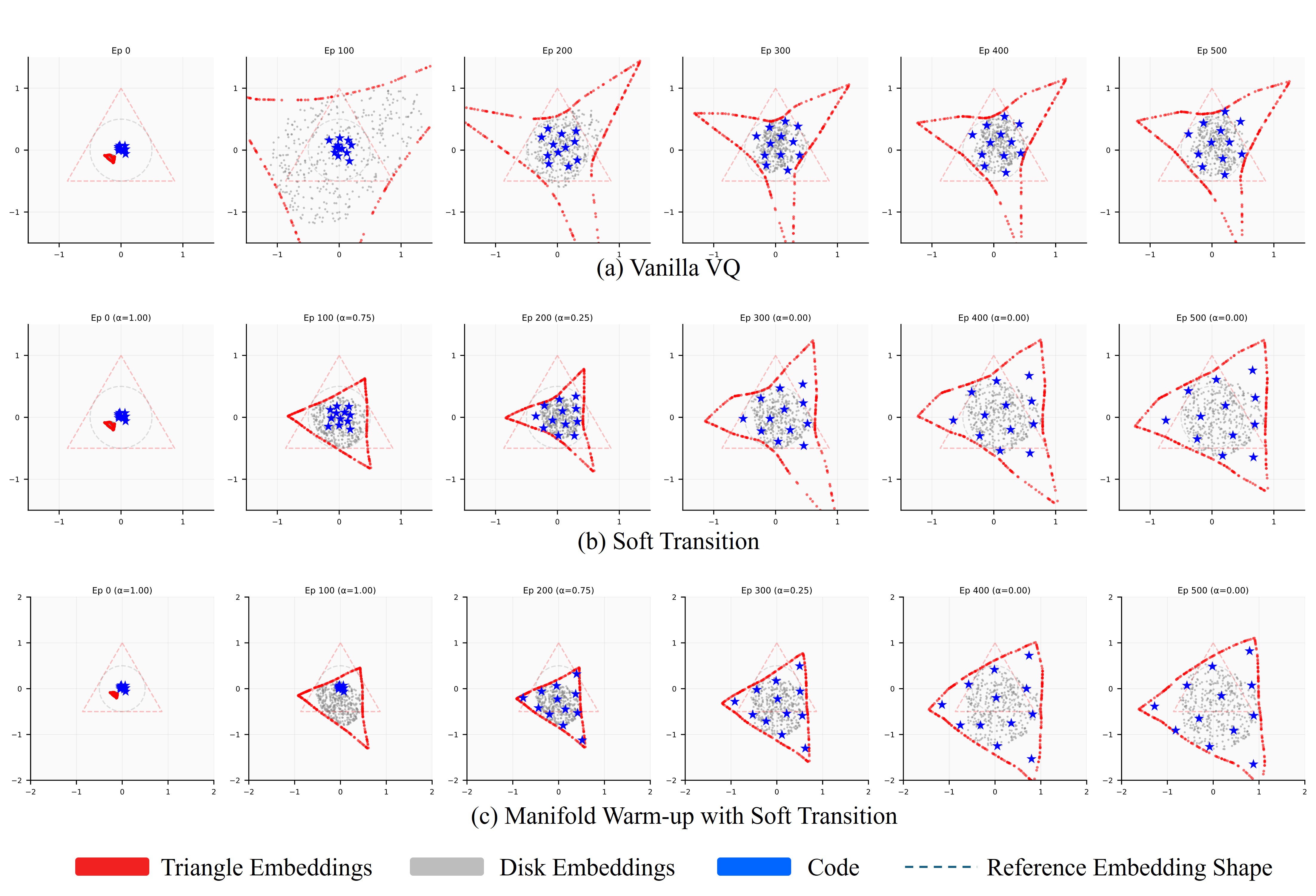}}
    \caption{\textbf{Comparison of Embedding and Codebook Dynamics during Training.} (a) Vanilla VQ: Inward-curved embedding edges signify grid mapping and an optimization deadlock, preventing full manifold coverage. (b) Soft Transition: Relaxes initial constraints to partially mitigate embedding shrinkage and improve codebook migration. (c) \texttt{ProVQ} (Ours): Manifold warm-up followed by soft transition achieves precise topological alignment, effectively resolving the deadlock.}
    \label{fig_synthetic_exp_manifold_change_v1}
  \end{center}
\end{figure*}

\vspace{-0.2cm}
\section{Phenomenon: Premature Discretization}
\label{sec:phenomenon}

To investigate the causes of premature discretization in VQ-VAEs, we design a controlled 2D synthetic diagnostic, which we call \textbf{TopoDisc} (Topology--Discretization Diagnostic). As shown in Figure \ref{fig_synthetic_exp_results_and_input} (a), this diagnostic consists of two distinct modes: \textit{Disk Data} ($n=400$), representing a dense central cluster, and \textit{Triangle Data} ($n=275$), forming a sharp boundary. This construction is specifically designed to expose discretization pathologies: the Disk component creates a centroid-attraction trap for the codebook, while the Triangle boundary makes grid-mapping artifacts and topological misalignment directly visible. Together, they form a minimal yet effective diagnostic tool for revealing whether a vector quantization method enforces discretization before the underlying manifold is properly discovered. Settings of TopoDisc can be adjusted for different discretization tasks and its codes are released in our github repo.

Our analysis reveals a performance gap in standard training. In Figure \ref{fig_synthetic_exp_results_and_input} (b), we observe that the reconstruction MSE for Triangle Data is substantially higher than for Disk Data ($0.0173$ vs. $0.0116$), indicating that vanilla VQ struggles to capture sharp geometric modes. However, as shown in Figure \ref{fig_synthetic_exp_results_and_input} (c), our proposed method, which disentangles the continuous and discrete effectively mitigate this gap.

\paragraph{Why Premature Discretization Occurs}
As visualized in Figure \ref{fig_synthetic_exp_manifold_change_v1} (a), vanilla VQ suffers from optimization stagnation almost immediately. When epoch is 0 $EP = 0$, the codebook (blue stars) is initialized without semantic information. By $Ep=300$, the encoder and codebook have entered a cycling co-adaptation phase: the encoder collapses its representation to minimize commitment loss toward the nearest (yet sub-optimal) codebook entries, and the code stagnates because of uninformative guidance from encoder. This leads to the encoder failing to unfold the whole data manifold correctly, leaving the Triangle Data mode poorly reconstructed even at $Ep=500$. This confirms that enforcing discretization before manifold warmup traps the system in a sub-optimal local minimum, a phenomenon we term \textit{Premature Discretization}.
To solve this problem, we proposed \texttt{ProVQ} and the results on synthentic dataset as show in Figure \ref{fig_synthetic_exp_results_and_input} and Figure \ref{fig_synthetic_exp_manifold_change_v1}.

\section{Progressive Vector Quantization (\texttt{ProVQ})}

Building upon our observation of the \textbf{co-adaptation deadlock}, we reformulate Vector Quantization (VQ) training as a Curriculum Learning task. Curriculum learning posits that models achieve superior convergence when introduced to tasks of increasing complexity. In the context of VQ-VAEs, the simultaneous training of a randomly initialized encoder $E_\theta$ and codebook $\mathcal{C} = \{e_i\}_{i=1}^K$ creates a ``complexity shock'' that often leads to sub-optimal local minima. To bypass this deadlock, we propose \textbf{Progressive Vector Quantization (\texttt{ProVQ})}, which decouples manifold warmup from latent discretization through a staged transition.

\subsection{Stage 1: Manifold Warmup (Easy Task)}
The initial phase of our curriculum focuses on \textbf{manifold warmup}. We utilize a standard continuous Autoencoder (AE) to capture the intrinsic global structure of the data distribution without the interference of quantization noise. By optimizing a standard reconstruction objective:
\begin{equation}
    \mathcal{L}_{\text{AE}} = \mathbb{E}_{x \sim p(x)} \left[ \|x - D_\phi(E_\theta(x))\|^2 \right],
\end{equation}
the encoder learns to map input data onto a continuous manifold that preserves essential features, such as sharp boundaries and disconnected modes. During this stage, the encoder unfolds complex data geometries, establishing a stable latent space that serves as a robust anchor for subsequent quantization. To bridge the gap between the continuous and discrete regimes, we initialize the codebook centroids by performing K-Means clustering on a batch of training embeddings.

\subsection{Stage 2: Scheduled Discretization (Hard Task)}
Once the manifold is established, the curriculum introduces the \textbf{discretization constraint} via a hybrid latent representation $\tilde{z}$ that smoothly interpolates between the continuous encoder output $z$ and its quantized counterpart $z_q$. In this \textbf{soft transition} stage, we define the quantized vector using the straight-through estimator (STE) as $z_q = \text{sg}[e_k] + z - \text{sg}[z]$, where $k = \arg\min_i \|z - e_i\|_2$. The soft transition is governed by a scheduling coefficient $\alpha(t)$:
\begin{equation}
    \tilde{z}(t) = \alpha(t) \cdot z + (1 - \alpha(t)) \cdot z_q.
\end{equation}

To facilitate a stable hand-off from continuous to discrete regimes, we employ a cosine-annealing scheduler for $\alpha(t)$ such that:
\begin{equation}
    \alpha(t) = 
    \begin{cases} 
    \frac{1}{2}\left[1 + \cos\left(\pi \frac{t}{T_{\text{trans}}}\right)\right], & 0 \le t < T_{\text{trans}} \\ 
    0, & t \ge T_{\text{trans}} 
    \end{cases}
\end{equation}
where $T_{\text{trans}}$ denotes the transition horizon. This schedule ensures the model is gradually ``weaned off'' continuous signals. Early in Stage 2, the encoder is allowed to migrate toward discrete representations via gradients from the reconstruction loss, facilitating a smooth adaptation to the discrete bottleneck without losing the underlying manifold structure.

The total training objective $\mathcal{L}_{\text{ProVQ}}$ is dynamically weighted to balance manifold preservation with quantization accuracy:
\begin{equation}
    \mathcal{L}_{\text{ProVQ}} = \mathcal{L}_{\text{recon}}(x, D_\phi(\tilde{z})) + \omega(t) \cdot \left( \mathcal{L}_{\text{VQ}} + \beta \mathcal{L}_{\text{commit}} \right),
\end{equation}
where $\mathcal{L}_{\text{VQ}} = \| \text{sg}[z] - z_q \|^2$ and $\mathcal{L}_{\text{commit}} = \| z - \text{sg}[z_q] \|^2$. The adaptive weight $\omega(t) = \lambda + (1 - \lambda) \cdot (1 - \alpha(t))$ gradually scales the influence of the quantization penalty. Here, $\lambda$ is used to control the initial coupling strength between the encoder and the codebook. 

By integrating the manifold warmup with the soft transition mechanism, we maintain gradient fluidity during the early training stages, preventing the encoder from being prematurely trapped in a local minimum because of grid mapping. Consequently, the final discretization becomes a targeted refinement of an already optimized latent partition rather than a constrained and noisy search.

\vspace{-0.2cm}
\section{Experiments}
\subsection{Experimental Setup}

\subsubsection{Synthetic Data}
To analyze the dynamics of \textit{premature discretization}, we design a 2D synthetic dataset featuring two distinct geometric components. One component is a high-density disk-shaped distribution intended to attract codebook entries toward the origin, thereby simulating the conditions that trigger sub-optimal grid mapping. Another part is triangular boundary dataset utilized to visualize latent distortions. Specifically, we assume the grid mapping is identified by the characteristic inward warping of the triangle's edges as the encoder prematurely collapses toward central centroids. Reconstruction quality is quantified using Mean Squared Error (MSE).

\subsubsection{Image Modality}
We evaluate \texttt{ProVQ} on ImageNet-100 and ImageNet-1K ($256 \times 256$) \cite{deng2009imagenet} for reconstruction and generation tasks. To ensure a stable and robust FID measurement on ImageNet-100, we build up the test set by uniformly sampling total 15,000 images from the training classes, plus a 5,000-image validation set. 

We quantify tokenizer quality using several standard metrics: reconstruction FID (rFID), PSNR, and SSIM for reconstruction fidelity, alongside Perplexity and average pairwise Euclidean distance to evaluate codebook utilization and diversity. Furthermore, generative performance is assessed through generation FID (gFID), Inception Score (IS), Precision, and Recall\cite{precisonandrecall} to provide a comprehensive view of the model's synthesis capabilities.

Our tokenizer follows the LlamaGen's VQGAN\cite{llamagen} configuration with a codebook size of 16,384 and a latent dimension of 8. Training includes a manifold warmup of 50,000 steps ($\sim$5 epochs) with a batch size of 128 and a loss weight $\lambda = 0.5$, followed by a 20,000-step cosine-scheduled soft transition. Generative models (LlamaGen-B and LlamaGen-L) are trained for 300 epochs following the original protocol. 

\subsubsection{Protein Modality}
In the biological domain, we utilize StructTokenBench \cite{yuan2025protein} as the benchmark for protein structure tokenization. Tokenizer effectiveness is assessed across 12 downstream tasks spanning 7 functional categories, such as Binding Interaction and Catalytic Site prediction. Additionally, we report token pair-wise euclidean distance and codebook utilization quantify the quality and diversity of codes.

Our implementation follows established training recipes for AminoAseed\cite{yuan2025protein} and Vanilla VQ tokenizers, both built upon the ESM3\cite{ESM3} architecture. We employ a manifold warmup of 20,000 steps with a batch size of 32 and $\lambda = 1.0$, followed by a 10,000-step soft transition period using a cosine scheduler.

\subsection{Image Reconstruction \& Generation}

To evaluate the practical efficacy of our proposed framework in natural image scenarios, we integrate the \texttt{ProVQ} tokenizer into the LlamaGen framework and conduct evaluations on the ImageNet-1K benchmark. This section analyzes both the reconstruction fidelity of the tokenizer and its downstream impact on generative performance across small and medium scale.

\definecolor{highlightblue}{RGB}{232, 242, 255}

\begin{table}[t]
\centering
\small
\caption{\textbf{Tokenizer performance on ImageNet-1K.} We compare the \texttt{ProVQ} with LlamaGen tokenizer across latent resolutions ($16\times16$ and $24\times24$).}
\setlength{\tabcolsep}{2pt}          
\renewcommand{\arraystretch}{1.1}   
\resizebox{\columnwidth}{!}{%
\begin{tabular}{ccccccc}
\toprule
Latent & Tokenizer & rFID$\downarrow$ & PSNR$\uparrow$ &SSIM $\uparrow$ & Perplexity$\uparrow$ & Euc dist.$\uparrow$ \\
\midrule
\multirow{2}{*}{16$\times$16}
  & LlamaGen & 2.19 & 20.79 & 0.675 & 8580.30 & 1.42 \\
  & \cellcolor{highlightblue} + \texttt{ProVQ}     & \cellcolor{highlightblue} \textbf{1.86} & \cellcolor{highlightblue} \textbf{20.92} & \cellcolor{highlightblue} \textbf{0.682} & \cellcolor{highlightblue} \textbf{8591.85} & \cellcolor{highlightblue} \textbf{6.49} \\
\cmidrule(lr){1-7}
\multirow{2}{*}{24$\times$24}
  & LlamaGen & 0.94 & 21.94 & 0.726 & 11487.83 & 1.42 \\
  & \cellcolor{highlightblue} + \texttt{ProVQ}    & \cellcolor{highlightblue} \textbf{0.81} & \cellcolor{highlightblue} \textbf{21.99} & \cellcolor{highlightblue} \textbf{0.729} & \cellcolor{highlightblue} \textbf{11551.56} & \cellcolor{highlightblue} \textbf{6.49}  \\
\bottomrule
\end{tabular}
}
\label{tab_IN1K_Reconstruction}
\vspace{-2.0em}
\end{table}

\definecolor{highlightblue}{RGB}{230, 240, 255}

\begin{table*}[ht]
\centering
\small          
\setlength{\tabcolsep}{8pt}
\caption{\textbf{Generative results on Imagenet1K($256 \times 256$).} Notably, integrating the \texttt{ProVQ} tokenizer into LlamaGen-B and LlamaGen-L architectures leads to consistent improvements in gFID and Recall. These results demonstrate that the enhanced reconstruction fidelity and discrete bottleneck by \texttt{ProVQ} translate to improved generative quality and more robust distribution coverage of ground truth.}
\begin{tabular}{llrrrrr}
\toprule
Type & Model & \#Para. & gFID$\downarrow$ & IS$\uparrow$ & Precision$\uparrow$ & Recall$\uparrow$ \\
\midrule
\multirow{4}{*}{Diffusion}
& ADM~\cite{adm}      & 554M & 10.94 & 101.0 & 0.69 & 0.63 \\
& CDM~\cite{cdm}               & --   &  4.88 & 158.7 & --   & --   \\
& LDM-4~\cite{ldm}       & 400M &  3.60 & 247.7 & --   & --   \\
& DiT-XL/2~\cite{dit}     & 675M &  2.27 & 278.2 & 0.83 & 0.57 \\
\midrule
\multirow{3}{*}{VAR}
& VAR-d16~\cite{tian2024visual}         & 310M & 3.30 & 274.4 & 0.84 & 0.51 \\
& VAR-d20~\cite{tian2024visual}         & 600M & 2.57 & 302.6 & 0.83 & 0.56 \\
& ImageFolder~\cite{li2024imagefolder}                      & 362M & 2.60 & 295.0 & 0.75 & 0.63 \\
\midrule
\multirow{9}{*}{AR}
& VQGAN~\cite{esser2021taming}         & 227M & 18.65 &  80.4 & 0.78 & 0.26 \\
& VQGAN~\cite{esser2021taming}          & 1.4B & 15.78 &  74.3 & --   & --   \\
& VQGAN-re~\cite{esser2021taming}       & 1.4B &  5.20 & 280.3 & --   & --   \\
& ViT-VQGAN~\cite{yu2021vector}        & 1.7B &  4.17 & 175.1 & --   & --   \\
& ViT-VQGAN-re~\cite{yu2021vector}      & 1.7B &  3.04 & 227.4 & --   & --   \\
& RQTran.~\cite{lee2022autoregressive}          & 3.8B &  7.55 & 134.0 & --   & --   \\
& RQTran.-re~\cite{lee2022autoregressive}       & 3.8B &  3.80 & 323.7 & --   & --   \\
& Open-MAGVIT2-AR-B~\cite{luo2024open}                  & 343M & 3.08 & 258.26 & 0.85 & 0.51 \\
& Open-MAGVIT2-AR-L~\cite{luo2024open}                   & 804M & 2.51 & 271.70 & 0.84 & 0.54 \\
\midrule
\multirow{4}{*}{AR}
& LlamaGen-B ~\cite{llamagen}     & 111M & 5.46 & 193.61 & 0.83 & 0.45 \\
& \cellcolor{highlightblue} \quad + \texttt{ProVQ} tokenizer (16 $\times$ 16)   
& \cellcolor{highlightblue} 111M 
& \cellcolor{highlightblue} 4.99 
& \cellcolor{highlightblue} 190.30 
& \cellcolor{highlightblue} 0.84 
& \cellcolor{highlightblue} 0.46 \\
& LlamaGen-L~\cite{llamagen}     & 343M & 3.80 & 248.28 & 0.83 & 0.51 \\
& \cellcolor{highlightblue} \quad + \texttt{ProVQ} tokenizer (16 $\times$ 16)   
& \cellcolor{highlightblue} 343M 
& \cellcolor{highlightblue} 3.15 
& \cellcolor{highlightblue} 235.51 
& \cellcolor{highlightblue} 0.82 
& \cellcolor{highlightblue} 0.54 \\
\bottomrule
\end{tabular}
\label{tab_generative_results}
\end{table*}

\subsubsection{Reconstruction Performance}

The reconstruction results summarized in Table \ref{tab_IN1K_Reconstruction} demonstrate that \texttt{ProVQ} consistently enhances reconstruction quality. At a $16 \times 16$ latent resolution, \texttt{ProVQ} improves rFID from $2.19$ to $1.86$ and increases PSNR from $20.79$ to $20.92$. Following the LlamaGen recipe, we also evaluate performance at an image resolution of $384 \times 384$ to obtain results at a $24 \times 24$ latent resolution. Similar gains are observed here, with rFID further reduced from $0.94$ to $0.81$. Beyond standard fidelity metrics, we observe an increase in codebook perplexity, rising from $8580.30$ to $8591.85$ at the $16 \times 16$ resolution, which indicates a more efficient and uniform utilization of codebook entries compared to the baseline.

A notable observation is the expansion of the average Euclidean distance between codes, which increases from $1.42$ to $6.49$. This increase suggests that \texttt{ProVQ} helps encoder embeddings to more broadly explore the latent manifold, potentially avoiding a collapse into a narrow  cluster. By alleviating the influence of sub-optimal grid mapping—a phenomenon where the encoder might otherwise over-simplify data structure—\texttt{ProVQ} allow the latent codes to better follow the encoder’s exploration of diverse modes. This improved coverage of the data distribution might help the tokenizer capture more nuanced semantic details, supporting high-fidelity image synthesis.

\subsection{Boosting Generative Image Models}

We further assess the impact of the \texttt{ProVQ} tokenizer on downstream generative tasks using LlamaGen-B and LlamaGen-L architectures. As shown in Table \ref{tab_generative_results}, the integration of \texttt{ProVQ} yields consistent improvements in generative quality across different model sizes. For the LlamaGen-B variant, \texttt{ProVQ} reduces the gFID from $5.46$ to $4.99$. For the larger LlamaGen-L model, the gFID improves from $3.80$ to $3.15$. Moreover, we observe the Recall consistently improve over .Such results point to a more robust capture of the ground-truth distribution, stemming from enhanced latent space utilization. Ultimately, the improved reconstruction fidelity afforded by \texttt{ProVQ} acts as a catalyst for superior generation, enhancing autoregressive model capacity through a more expressive and diverse discrete bottleneck.

Regarding the Inception Score (IS), we observe a marginal decrease, such as the shift from $248.28$ to $235.51$ for LlamaGen-L. We hypothesize that while \texttt{ProVQ} achieves a better overall match with the ground-truth data distribution, the smoother and more diverse latent space may reduce the over-fitting to specific class-discriminative features that the Inception-v3 classifier prioritizes.

\subsection{Protein Tokenization}

\definecolor{bestgreen}{RGB}{190, 230, 190} 
\definecolor{secondgreen}{RGB}{230, 245, 230} 

\begin{table*}[ht]
\centering
\caption{\textbf{Evaluation of ProVQ on StructTokenBench\cite{yuan2025protein}.} We compare the \texttt{ProVQ} progressive quantization strategy against several baselines: Van. VQ (Vanilla VQ based on ESM3) and AminoA. (AminoAseed) etc. Results demonstrate that the integration of \texttt{ProVQ} consistently improves both Vanilla VQ and AminoAseed. Notably, AminoA. + \texttt{ProVQ} achieves the highest average performance across all tasks, highlighted by a $9.69\%$ improvement(from $46.01 \%$ to $55.70 \%$) in structure property prediction.}
\label{tab:comparison_compact}
\footnotesize 
\renewcommand{\arraystretch}{1.2}
\setlength{\tabcolsep}{8pt} 

\resizebox{0.75\textwidth}{!}{%
\begin{tabular}{l|l|ccccc|ll}
\toprule
\multirow{3}{*}{\textbf{Task}} & \multirow{3}{*}{\textbf{Split}} & \multicolumn{5}{c|}{\textbf{Baselines}} & \multicolumn{2}{c}{\textbf{Ours}} \\ \cmidrule(lr){3-7} \cmidrule(lr){8-9}
 &  & FoldSeek & ProTokens & ESM3 & Van.VQ & AminoA. & Van.VQ & AminoA. \\ 
 &  &  &  &  &  &  & + \texttt{ProVQ} & + \texttt{ProVQ} \\ \midrule
\multicolumn{9}{l}{\textbf{Functional Site Prediction (AUROC\%)}} \\ \midrule
BindInt & Fold & \cellcolor{bestgreen}\textbf{53.18} & 44.66 & 44.30 & 47.25 & 47.11 & \cellcolor{secondgreen}\underline{48.95} & 48.28 \\
 & SupFam & 46.20 & 86.05 & 90.77 & 86.71 & 90.53 & \cellcolor{secondgreen}\underline{91.04} & \cellcolor{bestgreen}\textbf{91.55} \\
BindBio & Fold & 52.37 & 58.47 & 62.84 & 62.02 & \cellcolor{bestgreen}\textbf{65.73} & \cellcolor{secondgreen}\underline{65.36} & 63.90 \\
 & SupFam & 52.41 & 60.47 & 65.22 & 62.92 & \cellcolor{bestgreen}\textbf{68.30} & \cellcolor{secondgreen}\underline{67.55} & 66.76 \\
BindShake & Org & 53.40 & 59.82 & 66.10 & 67.04 & \cellcolor{bestgreen}\textbf{69.61} & 68.41 & \cellcolor{secondgreen}\underline{69.34} \\
CatInt & Fold & 53.43 & 58.16 & 61.09 & 58.89 & \cellcolor{secondgreen}\underline{62.19} & 61.62 & \cellcolor{bestgreen}\textbf{64.65} \\
 & SupFam & 51.41 & 83.85 & 89.82 & 85.00 & \cellcolor{secondgreen}\underline{91.91} & 90.94 & \cellcolor{bestgreen}\textbf{93.09} \\
CatBio & Fold & 56.37 & 56.14 & 65.33 & \cellcolor{bestgreen}\textbf{67.58} & \cellcolor{secondgreen}\underline{65.95} & 63.99 & 65.67 \\
 & SupFam & 53.78 & 64.05 & 74.65 & 70.92 & \cellcolor{secondgreen}\underline{87.59} & 84.42 & \cellcolor{bestgreen}\textbf{89.60} \\
Con & Fold & 49.26 & 56.23 & 55.22 & \cellcolor{secondgreen}\underline{56.98} & \cellcolor{bestgreen}\textbf{57.23} & 54.56 & 56.66 \\
 & SupFam & 51.39 & 74.33 & 80.53 & 74.60 & \cellcolor{bestgreen}\textbf{86.60} & 85.61 & \cellcolor{secondgreen}\underline{85.94} \\
Rep & Fold & 47.70 & \cellcolor{bestgreen}\textbf{77.25} & 74.70 & \cellcolor{secondgreen}\underline{75.99} & 74.97 & 74.65 & 75.32 \\
 & SupFam & 52.53 & 78.90 & 82.36 & 82.09 & \cellcolor{secondgreen}\underline{84.57} & 84.13 & \cellcolor{bestgreen}\textbf{86.04} \\
Ept & Fold & 54.52 & 54.69 & \cellcolor{secondgreen}\underline{63.69} & 59.28 & 62.16 & \cellcolor{bestgreen}\textbf{64.16} & 60.29 \\
 & SupFam & 50.56 & 67.52 & 61.97 & 67.24 & 72.02 & \cellcolor{bestgreen}\textbf{72.78} & \cellcolor{secondgreen}\underline{72.21} \\ \cmidrule(lr){1-9}
\textbf{Average} & & 51.90 & 65.37 & 69.24 & 68.30 & \cellcolor{secondgreen}\underline{72.43} & 71.88 & \cellcolor{bestgreen}\textbf{72.62} \\ \midrule
\multicolumn{9}{l}{\textbf{Physiochemical Property Prediction (Spearman's $\rho$\%)}} \\ \midrule
FlexRMSF & Fold & 15.35 & 13.81 & 44.53 & 44.22 & \cellcolor{secondgreen}\underline{44.63} & 43.87 & \cellcolor{bestgreen}\textbf{44.94} \\
 & SupFam & 11.99 & 7.62 & 39.68 & 39.08 & \cellcolor{secondgreen}\underline{40.99} & 40.10 & \cellcolor{bestgreen}\textbf{41.28} \\
FlexBFactor & Fold & 4.17 & 6.67 & \cellcolor{bestgreen}\textbf{23.60} & 22.32 & 21.30 & \cellcolor{secondgreen}\underline{23.34} & 22.97 \\
 & SupFam & 6.97 & 5.47 & \cellcolor{bestgreen}\textbf{25.80} & 23.73 & 21.76 & 24.59 & \cellcolor{secondgreen}\underline{24.61} \\
FlexNEQ & Fold & 5.71 & 12.98 & 45.08 & 35.95 & \cellcolor{secondgreen}\underline{49.64} & 48.01 & \cellcolor{bestgreen}\textbf{50.20} \\
 & SupFam & 2.60 & 12.50 & 45.43 & 35.61 & \cellcolor{bestgreen}\textbf{50.15} & 46.98 & \cellcolor{secondgreen}\underline{49.29} \\ \cmidrule(lr){1-9}
\textbf{Average} & & 7.80 & 9.84 & 37.35 & 33.49 & \cellcolor{secondgreen}\underline{38.08} & 37.82 & \cellcolor{bestgreen}\textbf{38.88} \\ \midrule
\multicolumn{9}{l}{\textbf{Structure Property Prediction (Macro F1\%)}} \\ \midrule
Homo & Fold & 11.57 & 5.84 & 30.02 & 18.17 & 29.87 & \cellcolor{secondgreen}\underline{31.94} & \cellcolor{bestgreen}\textbf{38.21} \\
 & SupFam & 4.67 & 6.17 & 24.89 & 22.10 & 38.38 & \cellcolor{secondgreen}\underline{38.54} & \cellcolor{bestgreen}\textbf{41.49} \\
 & Fam & 15.30 & 18.33 & 54.42 & 47.18 & \cellcolor{secondgreen}\underline{69.78} & 69.74 & \cellcolor{bestgreen}\textbf{87.39} \\ \cmidrule(lr){1-9}
\textbf{Average} & & 10.51 & 10.11 & 36.44 & 29.15 & 46.01 & \cellcolor{secondgreen}\underline{46.74} & \cellcolor{bestgreen}\textbf{55.70} \\ \bottomrule
\end{tabular}%
}
\label{tab_protein_effectiveness}
\end{table*}

\subsection{Evaluation on Protein Structure Modeling}

To further verify the generalization capabilities of \texttt{ProVQ} beyond the visual domain, we extend our evaluation to protein structure modeling using the PSTbench benchmark. Protein structures possess complex three-dimensional topologies that are highly sensitive to geometric fidelity, providing a rigorous testbed for our progressive quantization strategy. As detailed in Table \ref{tab_protein_effectiveness}, we compare \texttt{ProVQ} against several baselines including FoldSeek\cite{foldseek}, ProTokens\cite{protoken}, and ESM3-based tokenizers across 3 core aspects: functional site prediction, physiochemical property prediction, and homology detection.

In the Functional Site Prediction task, the integration of \texttt{ProVQ} yields consistent improvements in average AUROC. Specifically, while the vanilla VQ based on ESM3 achieves a mean of $68.30\%$, the addition of \texttt{ProVQ} increases the performance to $71.88\%$. When combined with the more advanced AminoAseed tokenizer, our method reaches a peak average of $72.62\%$, outperforming all baseline models. This trend is reflected in the Physiochemical Property Prediction task, where the \texttt{ProVQ}-enhanced AminoAseed achieves the highest mean score of $38.88\%$. These results suggest that the manifold warmup phase of our method allows the encoder to capture finer local geometric features and physiochemical nuances that are typically lost during the premature discretization of vanilla VQ-VAEs.

The most significant performance gain is observed in the Structure Property Prediction task.While the vanilla AminoAseed baseline achieves an average score of $46.01\%$, the integration of our progressive quantization strategy increases this metric to $55.70\%$. This improvement suggests that \texttt{ProVQ} effectively addresses potential sub-optimal quantization within the complex conformational manifolds of proteins. Given that remote homology detection relies heavily on the preservation of global structural motifs and long-range topological dependencies, avoiding the grid mapping trap allows discrete tokens to capture a more diverse set of structural modes. By better aligning the codebook with the encoder's latent space, \texttt{ProVQ} enhances the representative capacity of protein tokenizers for downstream biological discovery.
\vspace{-0.25em}

\begin{table}[h]
\centering
\renewcommand{\arraystretch}{1.2}
\caption{\textbf{Codebook analysis across StructTokenBench (CASP14 and CAMEO datasets).} Metrics include codebook utilization rate (UR\%), normalized perplexity, and average pairwise Euclidean distance (Euc. Distance). Results indicate that \texttt{ProVQ} consistently improves both the efficiency and diversity of the discrete latent space across different Vanilla VQ and AminoAseed.}
\vspace{0.25em}
\resizebox{\columnwidth}{!}{%
\begin{tabular}{l|cc|cc|c}
\toprule
\multirow{2}{*}{\textbf{Model}} & \multicolumn{2}{c|}{CASP14} & \multicolumn{2}{c|}{CAMEO} & \multirow{2}{*}{Euc. Distance $\uparrow$} \\ \cmidrule(lr){2-3} \cmidrule(lr){4-5}
 & UR\% $\uparrow$ & Perplexity $\uparrow$ & UR\% $\uparrow$ & Perplexity $\uparrow$ & \\ \midrule
VanillaVQ & 5.55 & 0.0339 & 5.60 & 0.0337 & 46.80 \\
\rowcolor{highlightblue} \quad + \texttt{ProVQ} & 41.40 & 0.2985 & 43.19 & 0.3006 & \textbf{69.68} \\
AminoAseed & 64.45 & 0.4946 & 68.87 & 0.5119 & 42.71 \\ 
\rowcolor{highlightblue} \quad + \texttt{ProVQ} & \textbf{78.36} & \textbf{0.6021} & \textbf{85.36} & \textbf{0.6276} & 43.67 \\
\bottomrule
\end{tabular}
}
\label{tab_codebook_analysis_pstbench}
\end{table}
\vspace{1.0em}

As illustrated in Table \ref{tab_codebook_analysis_pstbench}, \texttt{ProVQ} acts as a robust regularizer for the discrete latent space by decoupling manifold warmup from discretization. This strategic separation prevents the encoder from being prematurely constrained by a rigid codebook, instead facilitating synchronized co-adaptation between the encoder's embeddings and codebook updates.The resulting improvements in normalized perplexity and average pairwise Euclidean distance across the CASP14 and CAMEO benchmarks underscore \texttt{ProVQ}'s ability to preserve high representation diversity. Specifically, for the VanillaVQ baseline, the Euc. Distance increases from 46.80 to 69.68, indicating a more expansive coverage of the latent space. Such diversity is fundamental for effectively capturing the vast representational variety inherent in complex protein structures.

\vspace{-0.2cm}
\section{Ablation Study}

\definecolor{lightishblue}{RGB}{235, 245, 255}  
\definecolor{strongerblue}{RGB}{210, 230, 255}  

\begin{table}[h]
    \centering
    \caption{\textbf{Ablation of Soft Transition and Manifold warmup on ImageNet-100 ($256 \times 256$).} The \textit{best} indicates a manifold warmup phase where the autoencoder (AE) reaches its peak validation rFID. The \textit{overfit} label denotes an extended warmup duration where the AE exhibits overfitting on the rFID.}
    \label{tab_ablation_IN100_soft_warmup}
    \small
    
    \renewcommand{\arraystretch}{1.3}

    \resizebox{\columnwidth}{!}{%
    \begin{tabular}{llcccc} 
        \toprule
        \textbf{Base Model} & \textbf{Enhancement} & \textbf{rFID}$\downarrow$ & \textbf{PSNR}$\uparrow$ & \textbf{SSIM}$\uparrow$ & \textbf{Perplexity}$\uparrow$ \\
        \midrule
        SimVQ & --- & 4.08 & 20.33 & 0.614 & 8,157.35 \\
        \rowcolor[HTML]{EBF5FB} 
        SimVQ & + Soft Transition & 3.39 & 20.53 & 0.628 & 8,171.83 \\
        \midrule
        Vanilla & --- & 3.81 & 20.64 & 0.629 & 7,123.79 \\
        \rowcolor[HTML]{EBF5FB} 
        Vanilla & + Soft Transition & 3.49 & 20.45 & 0.624 & \textbf{8,530.51} \\
        \rowcolor[HTML]{EBF5FB} 
        Vanilla & + Manifold Warmup \textit{(best)} & 3.66 & 20.62 & 0.636 & 8,442.90 \\
        \rowcolor[HTML]{EBF5FB} 
        Vanilla & + Manifold Warmup \textit{(overfit)} & 3.64 & 20.47 & 0.628 & 8,460.61 \\
        \midrule
        \rowcolor[HTML]{D6EAF8} 
        \textbf{Vanilla} & \textbf{+ Both (\texttt{ProVQ})} & \textbf{3.33} & \textbf{20.75} & \textbf{0.640} & 8,519.23 \\
        \bottomrule
    \end{tabular}
    }
\end{table}

We conduct extensive ablation studies on the ImageNet-100 dataset to isolate the contributions of each proposed component. To ensure a fair and consistent comparison, all experiments utilize the LlamaGen tokenizer architecture and are trained for the same 200 epochs to ensure convergence. 

\paragraph{Manifold Warmup} As shown in Table \ref{tab:ablation_soft}, the manifold warmup improved vanilla VQ from $3.81$ to $3.66$. Moreover, there is no significant difference in final tokenizer performance whether the autoencoder is warmed up to its best validation rFID at approximately 30 epochs or allowed to train further to a state of overfitting at 40 epochs. This observation suggests that once the latent manifold is sufficiently established, the subsequent transition mechanism is robust enough to handle minor variations in the continuous starting point.

\paragraph{Soft Transition} The soft transition mechanism further enhances reconstruction fidelity by providing a gradual shift into the discrete bottleneck. The impact of this transition is most pronounced in the SimVQ configuration, where the rFID is reduced from $4.08$ to $3.39$. We hypothesize that the relatively poor performance of the baseline SimVQ\cite{simvq} is due to an excessively rapid codebook optimization process which triggers an early grid mapping trap. Without a soft transition period, codebook entries converge prematurely, effectively locking the encoder into a sub-optimal state and preventing it from adequately exploring the embedding space. By employing our soft transition strategy with manifold warmup, we mitigate the premature coupling, allowing the encoder to develop more expressive and diverse representations before full discretization is enforced. Ultimately, the full \texttt{ProVQ} strategy achieves the best overall performance with an rFID of $3.33$ and a PSNR of $20.75$, confirming that a progressive approach to quantization is an effective method to boost performance.

\begin{table}[htbp]
    \centering
    \caption{\textbf{Ablation for scheduler setting of soft transition.} The cosine means cosine-annealing scheduler while hard scheduler maintains the $\alpha= 1$. Results show that gradual reduction of $alpha$ provides a soft landing and consequently improves reconstruction performance.  }
    \label{tab:ablation_soft}
    \resizebox{0.85\columnwidth}{!}{%
    \begin{tabular}{ccccc}
        \toprule
        Scheduler & rFID$\downarrow$ & PSNR$\uparrow$ & SSIM$\uparrow$ & Perplexity$\uparrow$ \\
        \midrule
        \rowcolor{highlightblue}
        cosine & \textbf{3.23} & 20.48 & 0.627 & 8518.33 \\
        hard   & 3.40 & \textbf{20.49} & \textbf{0.633} & \textbf{8532.74} \\
        \bottomrule
    \end{tabular}
    }
    \label{tab_ablation_soft}
\end{table}

\paragraph{Scheduler} As shown in Table \ref{tab:ablation_soft}, we compare our \textit{cosine-annealing scheduler} against a \textit{hard scheduler}, where the transition coefficient $\alpha$ remains at 1.0 throughout the transition phase before switching abruptly. The results show that the hard scheduler performing at $3.40$ rFID is notably inferior to the cosine scheduler at $3.23$. This gap underscores the importance of a smooth annealing process, a gradual reduction of $\alpha$ provides a soft landing that allows the encoder and codebook to maintain alignment as the latent space transitions from a continuous manifold to a discrete set of points.
\vspace{-0.2cm}
\section{Conclusion}
In this paper, we introduce Progressive Vector Quantization (\texttt{ProVQ}), a curriculum-inspired training strategy designed to overcome the fundamental co-adaptation deadlock inherent in standard Vector Quantization. 

Through extensive empirical validation, we have shown that \texttt{ProVQ} effectively prevents the grid mapping trap. Our results on ImageNet-1K and ImageNet-100 demonstrate that \texttt{ProVQ} significantly enhances both reconstruction fidelity and generative performance.
Beyond general vision tasks, \texttt{ProVQ} proves highly effective for modeling complex biological data. Most notably, \texttt{ProVQ} establishes a new performance ceiling for protein structure tokenization on the StrutTokenBench leaderboard, underscoring its versatility in capturing the precise structural modes required for biological sequence modeling. Ultimately, \texttt{ProVQ} provides a stable and robust framework for bridging continuous signals with discrete symbolic processing across diverse modalities.

\section*{Impact Statement}
This paper presents work whose goal is to advance the field of tokenization strategy. There are many potential societal consequences of our work, none which we feel must be specifically highlighted here.

\bibliography{main}
\bibliographystyle{icml2026}

\newpage
\appendix

\onecolumn

\end{document}